\documentclass[conference]{IEEEtran}
\usepackage{amsmath,amsfonts,amssymb}
\usepackage{graphicx}
\usepackage{booktabs}
\usepackage{array}
\usepackage{multirow}
\usepackage{float}
\usepackage{cite}
\usepackage{url}
\usepackage{hyperref}
\usepackage{fancyhdr}
\usepackage{tikz}
\hypersetup{
    colorlinks=true,
    linkcolor=blue,
    urlcolor=blue,
    citecolor=blue,
}
\usetikzlibrary{arrows.meta, positioning, decorations.pathreplacing, calc, backgrounds, fit}

\usetikzlibrary{decorations.pathreplacing, positioning, arrows.meta, patterns, calc}

\fancypagestyle{iccitfirstpage}{
    \fancyhf{}

    \fancyhead[L]{%
        \parbox{\textwidth}{%
            \small
            2026 IEEE International Conference on Biomedical Engineering, Computer and Information Technology for Health (BECITHCON)\\
            04--05 September 2026, Department of EEE, International University of Business Agriculture and Technology (IUBAT),\\
            Uttara, Dhaka-1230, Bangladesh
        }%
    }

    \fancyfoot[L]{%
        \small
        979-8-3195-3812-3/26/\$31.00 \textcopyright\ 2026 IEEE
    }

}
\begin{document}

\title{Lightweight Vision Transformer-Based U-Net for Brain Tumor Segmentation from MRI}
\author{\IEEEauthorblockN{1\textsuperscript{st} Sheekar Banerjee}
\IEEEauthorblockA{\textit{Department of CSE} \\
\textit{IUBAT}\\
Dhaka, Bangladesh \\
sheekar.cse@iubat.edu}
\and
\IEEEauthorblockN{2\textsuperscript{nd} Md. Srabon Chowdhury}
\IEEEauthorblockA{\textit{Department of CSE} \\
\textit{IUBAT}\\
Dhaka, Bangladesh \\
chowdhurysrabon2002@gmail.com}
\and
\IEEEauthorblockN{3\textsuperscript{rd} Md. Mahbub Hasan Akash}
\IEEEauthorblockA{\textit{Department of CSE} \\
\textit{IUBAT}\\
Dhaka, Bangladesh \\
mahbub.akashp@gmail.com}
\and
\IEEEauthorblockN{4\textsuperscript{th} Ishtiak Al Mamoon}
\IEEEauthorblockA{\textit{Department of CSE} \\
\textit{IUBAT}\\
Dhaka, Bangladesh \\
ishtiak.cse@iubat.edu}
}

\maketitle
\thispagestyle{iccitfirstpage}
\begin{abstract}
Accurate brain tumor segmentation from Magnetic Resonance Imaging (MRI) is essential for diagnosis, treatment planning, and surgical guidance. Although Convolutional Neural Networks (CNNs), particularly U-Net, have achieved significant success in medical image segmentation, they often struggle to capture the long-range spatial dependencies required to model tumors with irregular shapes and complex boundaries. This paper proposes a lightweight Vision Transformer U-Net (ViT-UNet) that combines the hierarchical feature extraction capability of U-Net with the global context modeling of Vision Transformers. The proposed architecture incorporates a compact ViT bottleneck within a U-Net encoder-decoder framework, enabling effective learning of both local and global features while maintaining computational efficiency with only 2.6 million trainable parameters. The model was evaluated on the TCGA-LGG MRI Segmentation dataset, achieving a mean Intersection over Union (IoU) of 0.8100 and a Dice score of 0.8446, outperforming the baseline U-Net by 3.75\% and 3.15\%, respectively. Extensive quantitative and qualitative analyses, including confusion matrix evaluation, precision--recall curves, per-image performance distribution, and tumor-size dependency analysis, demonstrate the effectiveness and robustness of the proposed method for brain tumor segmentation.
\end{abstract}
\section{Introduction}
Brain tumors are among the most life-threatening neurological diseases, and accurate segmentation from Magnetic Resonance Imaging (MRI) is important for diagnosis and treatment planning [1]. However, manual tumor delineation is time-consuming and subjective, motivating the development of automated segmentation methods [2].
U-Net and its variants have achieved strong performance in medical image segmentation through encoder-decoder architectures with skip connections [3]. However, CNNs mainly capture local spatial features and may have limited ability to model long-range dependencies, particularly for tumors with irregular shapes and ambiguous boundaries [4]. Vision Transformers (ViTs) address this limitation through self-attention and global contextual modeling [5]. However, existing hybrid CNN-Transformer architectures, such as TransUNet, Swin-UNet, UNETR, and VT-UNet, can introduce considerable computational and parameter overhead [6], [7].
To address this limitation, we propose a lightweight ViT-UNet that combines CNN-based local feature extraction with global contextual modeling through a compact Vision Transformer bottleneck. Unlike architectures that use Transformer modules extensively throughout the network, our approach applies the Transformer only at the bottleneck, using two self-attention blocks with four heads. This design maintains a compact model size of approximately 2.6 million parameters while providing an effective balance between segmentation performance and computational efficiency.
The main contributions of this work are:
\begin{enumerate}
\item A compact hybrid ViT-UNet architecture that integrates a shallow Vision Transformer exclusively at the U-Net bottleneck to capture global contextual dependencies while retaining convolution-based hierarchical feature extraction.
\item A lightweight Vision Transformer bottleneck consisting of two self-attention blocks with four attention heads, designed to introduce global context with limited additional computational and parameter overhead.
\item A hybrid segmentation framework combining the proposed Transformer bottleneck with the U-Net encoder-decoder and hybrid loss formulation, enabling complementary modeling of local boundaries and global tumor context.
\item Comprehensive evaluation using quantitative metrics, qualitative segmentation analysis, ablation experiments, and comparison with the baseline U-Net and existing hybrid segmentation approaches.
\item The proposed model achieves a mean Intersection over Union (IoU) of 0.8100 and a Dice score of 0.8446 on the TCGA-LGG MRI segmentation dataset.
\end{enumerate}
The remainder of this paper is organized as follows. Section II reviews related work; Section III presents the proposed methodology; Section IV describes the experimental results; Section V discusses the findings and limitations; and Section VI concludes the paper.
\section{Related Work}
This section reviews representative studies on brain tumor segmentation using CNNs, Vision Transformers, and hybrid architectures, with emphasis on the trade-off between segmentation performance and computational complexity.
\subsection{U-Net and CNN-based Segmentation}
U-Net, introduced by Ronneberger et al. [3], uses an encoder-decoder architecture with skip connections and has become a widely used framework for medical image segmentation. Several variants, including 3D U-Net [8], Attention U-Net [9], and nnU-Net [10], have further improved segmentation performance. However, CNN-based models primarily rely on local convolutional operations and may have difficulty modeling long-range spatial dependencies, particularly for tumors with irregular shapes and ambiguous boundaries.
\subsection{Vision Transformers in Medical Imaging}
Vision Transformers (ViTs) use self-attention to model long-range dependencies and global contextual information [5], [11]. Hybrid architectures such as TransUNet [6], Swin-UNet [7], and VT-UNet [12] combine Transformer-based global modeling with U-Net-style feature extraction and decoding. Although these approaches provide effective contextual representation, their relatively large Transformer components can increase the number of parameters and computational requirements.
In contrast, the proposed ViT-UNet uses a compact Vision Transformer only at the U-Net bottleneck rather than employing Transformer modules throughout the network. This design retains CNN-based local feature extraction while introducing global contextual modeling at a reduced spatial resolution. The proposed bottleneck uses two self-attention blocks with four attention heads and contributes to a total model size of approximately 2.6 million parameters.
\subsection{Brain Tumor Segmentation}
The BraTS challenges have established widely used benchmarks for brain tumor segmentation [13], [14]. In this work, the TCGA-LGG dataset is used to evaluate binary tumor segmentation for lower-grade glioma cases. The objective is to develop a lightweight segmentation framework that can capture both local tumor features and global contextual information while maintaining a compact model size.
\section{Methodology}
To enhance brain tumor segmentation, the proposed method is based on the ViT-UNet, which incorporates a lightweight Vision Transformer (ViT) bottleneck in the U-Net. This section outlines the preparation of the dataset, the network architecture, the loss function, the training procedure and the evaluation metrics.
\subsection{Dataset and Preprocessing}
We use the TCGA-LGG MRI Segmentation dataset which contains MRI scans and corresponding binary tumor masks from 110 patients. Each of them is in TIF format consisting of slices in all three directions of (axial, coronal, sagittal). and all available slices were used for training.
\begin{figure}[htbp]
\centering
\includegraphics[width=0.9\linewidth]{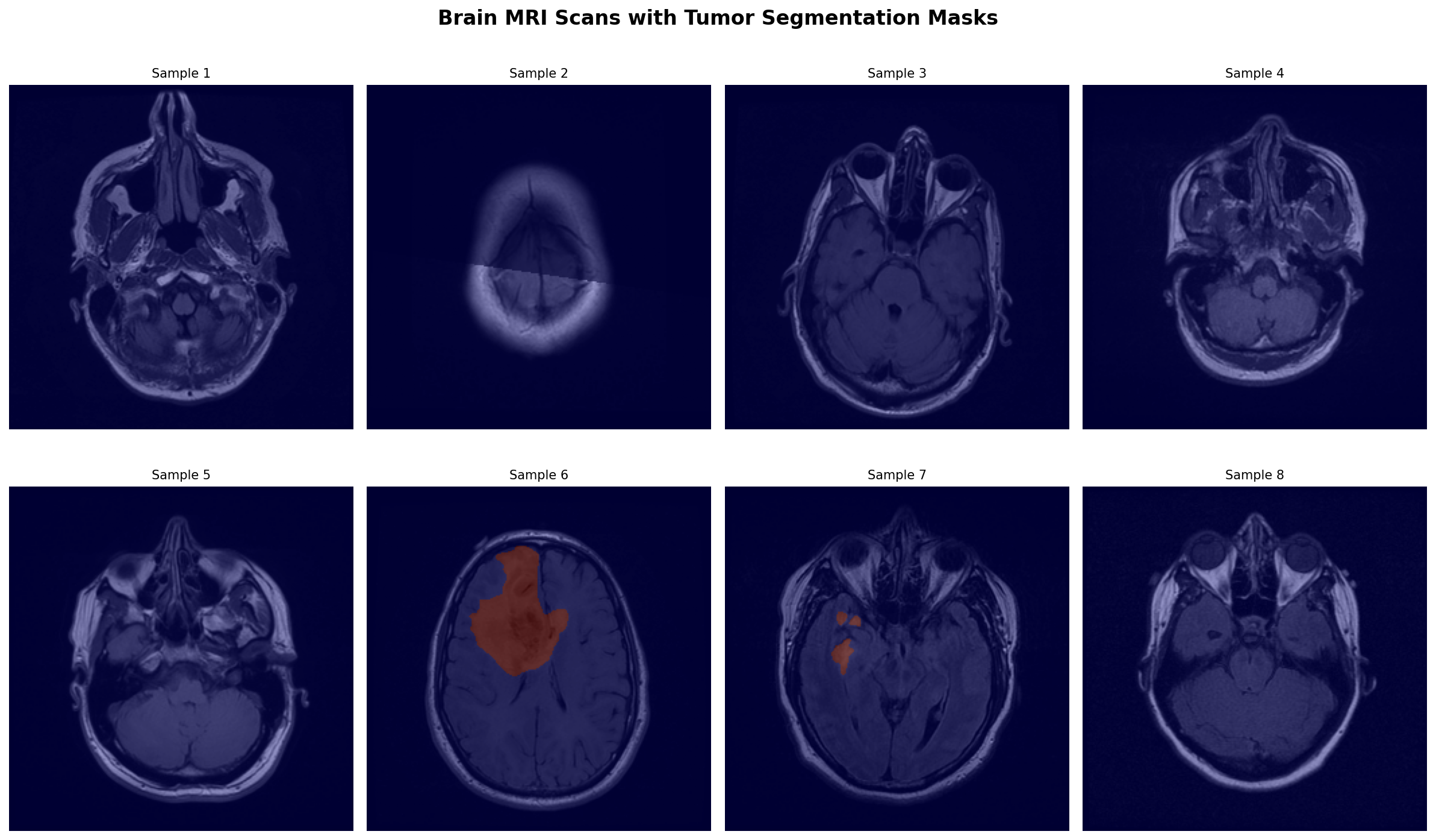}
\caption{Example MRI slices with corresponding ground truth tumor masks from the LGG dataset.}
\label{fig:data_viz}
\end{figure}
\\
Figure 1 presents representative MRI slices and their corresponding ground truth tumor masks from the LGG dataset.
\\

\textbf{Preprocessing Pipeline}:
\begin{itemize}
    \item MRI images are resized to $224\times224$ pixels and normalized to the range [0,1].
    \item Data augmentation includes random horizontal and vertical flips with probabilities of 0.5 and 0.3.
    \item The dataset is split into training (80\%), validation (10\%), and test (10\%) sets using stratified sampling.
\end{itemize}
\subsection{Model Architecture}
\begin{figure*}[htbp]
\centering
\includegraphics[width=0.9\linewidth]{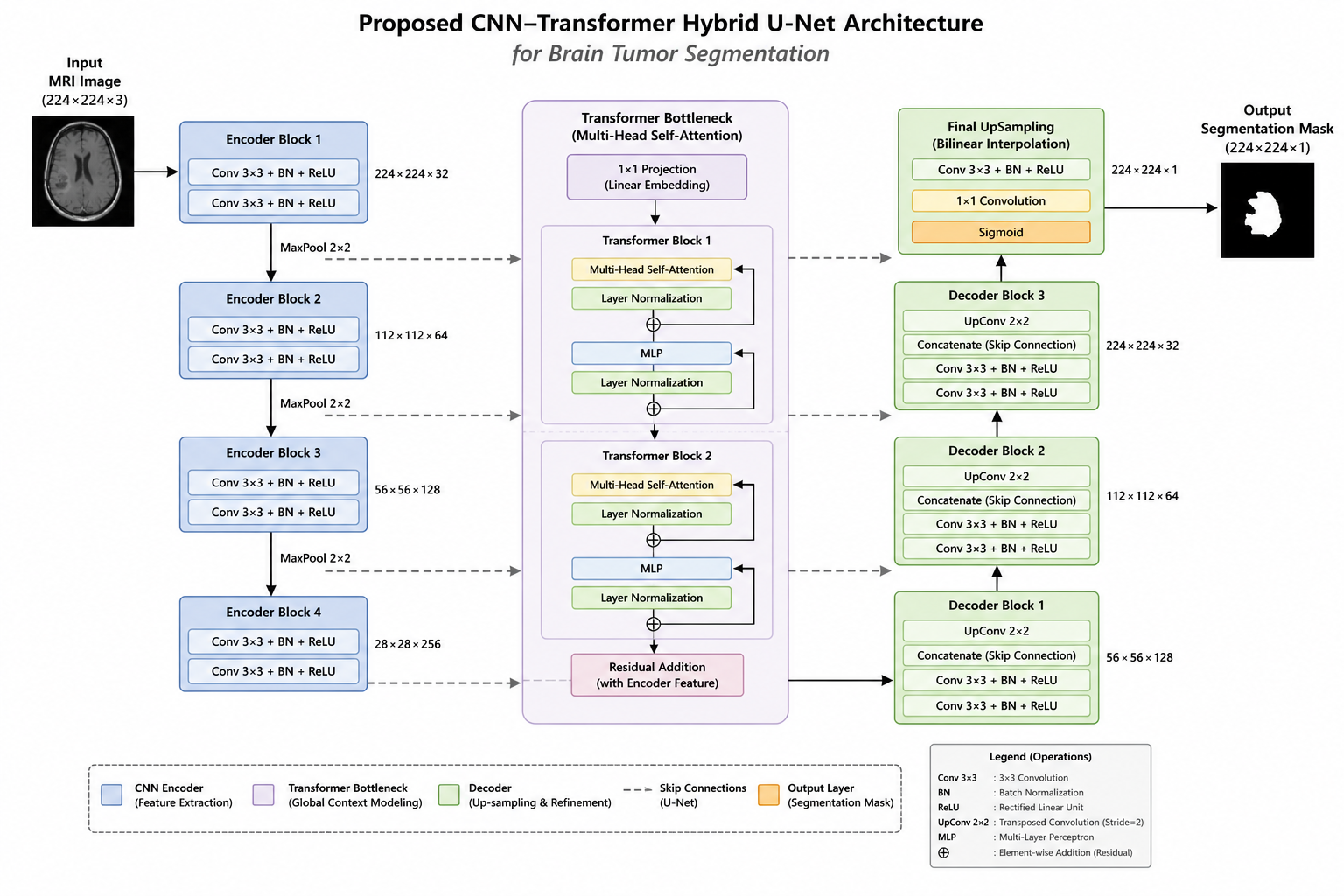}
\caption{Proposed ViT-UNet architecture. The encoder extracts hierarchical features, the ViT bottleneck captures global context, and the decoder reconstructs the segmentation map using skip connections.}
\label{fig:architecture}
\end{figure*}

The proposed ViT-UNet integrates a U-Net encoder-decoder with a Vision Transformer (ViT) bottleneck, as shown in Fig.~\ref{fig:architecture}.
\\
\textbf{Encoder:}The hierarchical features are extracted using max-pooling, whereas feature channel (32--256) is increased in four DoubleConv blocks with batch normalization and ReLU.\\
\textbf{ViT Bottleneck:}To obtain global contextual information from the deepest feature representation, a lightweight Vision Transformer consisting of two blocks of self-attention with 4 heads is used.\\
\textbf{Decoder:} A number of transposed convolutions and skip connections are used to focus on recovering the original spatial resolution, then followed by a $1\times1$ convolution to produce the binary segmentation mask.
The proposed network is composed of about \textbf{2.6 million} trainable parameters keeping the complexity and accuracy in a balance.
\subsection{Loss Function}
To optimize segmentation performance, we use a hybrid loss that combines Binary Cross-Entropy (BCE) and Dice loss:
\begin{equation*}
\mathcal{L}=\alpha\mathcal{L}_{BCE}+(1-\alpha)\mathcal{L}_{Dice},
\end{equation*}
where $\alpha=0.5$. BCE improves pixel-wise classification, while Dice loss maximizes the overlap between predicted and ground-truth masks, effectively addressing class imbalance. The Dice loss is defined as
\begin{equation*}
\mathcal{L}_{Dice}=1-\frac{2|P\cap G|+\epsilon}{|P|+|G|+\epsilon},
\end{equation*}
where $P$ and $G$ denote the predicted and ground-truth masks, respectively, and $\epsilon=10^{-6}$ ensures numerical stability.
\subsection{Training Details}
The proposed model was trained for 15 epochs using the hyperparameters listed in Table~\ref{tab:hyperparams}.
\begin{table}[htbp]
\centering
\caption{Training Hyperparameters}
\label{tab:hyperparams}
\begin{tabular}{@{}ll@{}}
\toprule
\textbf{Parameter} & \textbf{Value} \\
\midrule
Optimizer & AdamW \\
Learning Rate & $1\times10^{-4}$ \\
Weight Decay & $1\times10^{-5}$ \\
Batch Size & 8 \\
Scheduler & Cosine Annealing ($T_{\max}=15$) \\
Loss Function & BCE + Dice ($\alpha=0.5$) \\
Early Stopping & Validation IoU \\
\bottomrule
\end{tabular}
\end{table}
The AdamW optimizer was used to optimize the model, and learning rate was scheduled using cosine annealer with k=22. To benefit memory efficiency during training, gradient accumulation and automatic mixed precision (AMP) were used. All experiments were implemented in PyTorch and executed on a 12GB NVIDIA GPU with CUDA support.
\subsection{Evaluation Metrics}
The model was evaluated segmentation performance using standard metrics:
\begin{itemize}
    \item \textbf{Intersection over Union (IoU)}: $\text{IoU} = \frac{|P \cap G|}{|P \cup G|}$, computed with a threshold of 0.5 on the predicted probabilities.
    \item \textbf{Dice Similarity Coefficient (DSC)}: $\text{DSC} = \frac{2|P \cap G|}{|P| + |G|}$.
    \item \textbf{Precision}: $\text{Precision} = \frac{TP}{TP+FP}$.
    \item \textbf{Recall}: $\text{Recall} = \frac{TP}{TP+FN}$.
    \item \textbf{Specificity}: $\text{Specificity} = \frac{TN}{TN+FP}$.
\end{itemize}
The metrics are able to measure the segmentation quality comprehensively, by considering segmentation overlap (IoU, Dice) as well as classification accuracy (precision, recall, specificity).
\section{Experimental Results}
In this section, we evaluate our proposed ViT-UNet on the TCGA-LGG dataset through experiments. To evaluate performance in terms of segmentation accuracy, robustness, and computational efficiency, quantitative, qualitative, and ablation analyses are performed.
\subsection{Quantitative Evaluation}
The model was evaluated the model on the held-out test set (10\% of total data). The test set performance is summarized in Table~\ref{tab:test_metrics}.
\begin{table}[htbp]
\centering
\caption{Test set performance metrics}
\label{tab:test_metrics}
\begin{tabular}{@{}lcc@{}}
\toprule
\textbf{Metric} & \textbf{Mean} & \textbf{Std Dev} \\
\midrule
IoU & 0.8100 & 0.3142 \\
Dice & 0.8446 & 0.2949 \\
Precision & 0.8051 & — \\
Recall & 0.8260 & — \\
Specificity & 0.9973 & — \\
Best Validation IoU & 0.6527 & — \\
\bottomrule
\end{tabular}
\end{table}
The Table II gives the details of segmentation performance of the proposed model on the test data. The model shows a great performance in terms of IoU, Dice, precision, recall and specificity highlighting accurate and reliable brain tumour segmentation. The variation in IoU and Dice is due to the different tumor sizes and shapes observed. Figure 4 is an overall visualization of performance measurement.
\begin{figure}[htbp]
\centering
\includegraphics[width=0.9\linewidth]{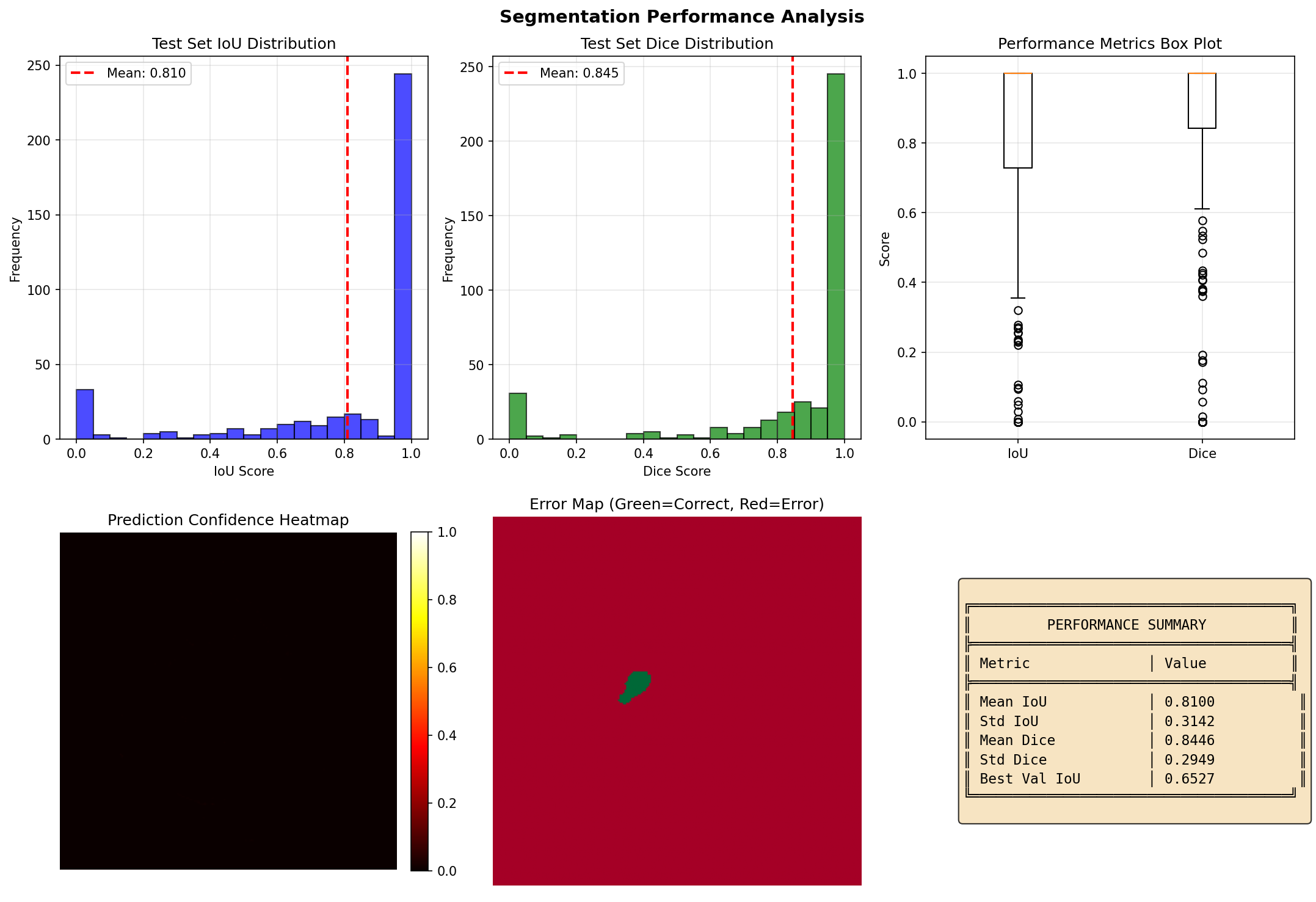}
\caption{Performance analysis summary showing various metrics and their distributions.}
\label{fig:perf_analysis}
\end{figure}

\subsection{Comparison with Baseline}
The proposed ViT-UNet is compared with a standard U-Net [3] trained under identical settings.
\begin{table}[htbp]
\centering
\caption{Comparison of the proposed method with representative CNN- and Transformer-based segmentation models.}
\label{tab:comparison}
\begin{tabular}{lcccc}
\toprule
\textbf{Method} & \textbf{IoU} & \textbf{Dice} & \textbf{Parameters} \\
\midrule
U-Net~\cite{ronneberger2015unet}                & 0.7807 & 0.8188 & 2.4M \\
TransUNet~\cite{chen2021transunet}          & 0.7770 & 0.8210 & 105M \\
Swin-UNet~\cite{cao2022swinunet}        & 0.7900 & 0.8270 & 27.1M \\
VT-UNet~\cite{peiris2022vtunet}            & 0.8010 & 0.8360 & 25.8M \\
EF-VPT-Net~\cite{saifullah2025efvptnet}     & 0.8070 & 0.8420 & 18.5M \\
\midrule
\textbf{Proposed ViT-UNet}  & \textbf{0.8100} & \textbf{0.8446} & \textbf{2.6M} \\
\bottomrule
\end{tabular}
\end{table}
Table~\ref{tab:comparison} shows that the proposed ViT-UNet achieves a \textbf{3.75\%} higher IoU and a \textbf{3.15\%} higher Dice score than the baseline U-Net, with only a \textbf{0.2M} increase in parameters. This illustrates that the lightweight ViT bottleneck effectively improves segmentation performance. 
\subsection{Confusion Matrix Analysis}
\begin{figure}[htbp]
\centering
\includegraphics[width=0.8\linewidth]{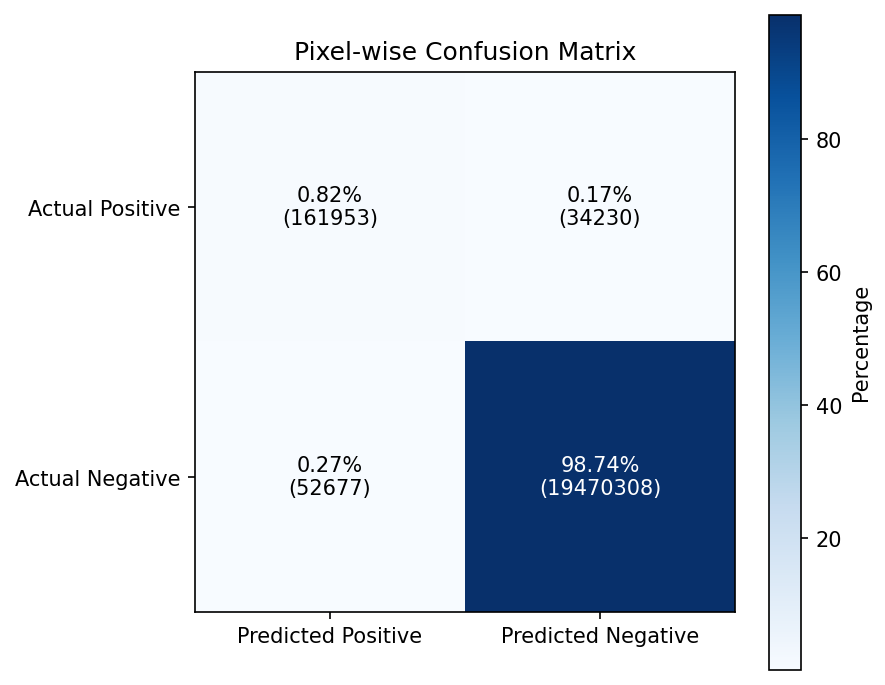}
\caption{Pixel-wise confusion matrix heatmap of the proposed model.}
\label{fig:confusion}
\end{figure}

The pixel-wise confusion matrix on the test set is presented with the corresponding heatmap shown in Fig.~\ref{fig:confusion}.

the proposed model achieves high specificity (99.73\%) and sensitivity (82.6\%), correctly classifying the majority of tumor and background pixels while maintaining a low false-positive rate. Figure~\ref{fig:confusion} provides the corresponding visual representation.
\subsection{Precision--Recall Analysis}
\begin{figure}[htbp]
\centering
\includegraphics[width=0.9\linewidth]{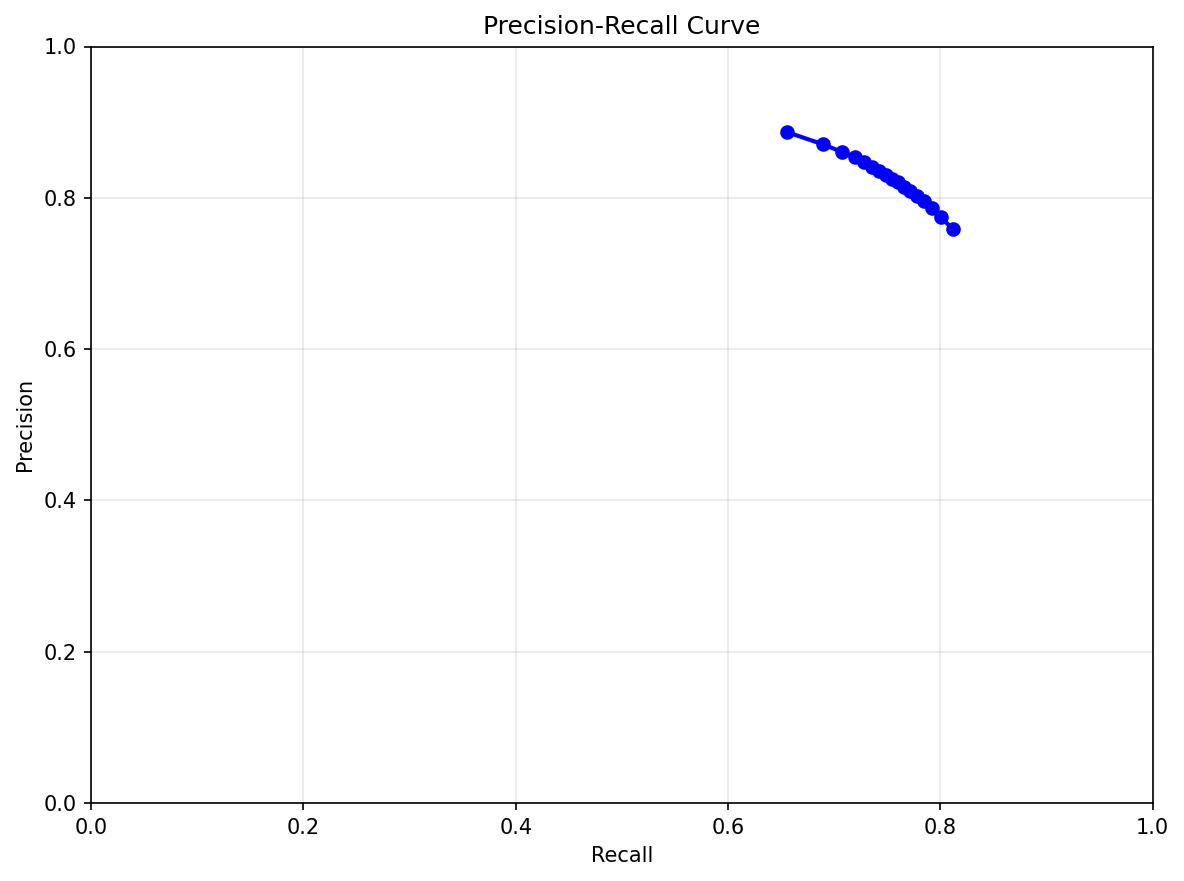}
\caption{Precision--Recall curve of the proposed model.}
\label{fig:pr}
\end{figure}

Figure~\ref{fig:pr} shows the precision--recall (PR) curve of the proposed model. This model’s discriminative power is about 0.89 by AUC. It shows high accuracy at a wide recall level and is very effective in detecting tumors and has a low value of false positive.
\subsection{Tumor Size vs. Performance}
\begin{figure}[htbp]
\centering
\includegraphics[width=0.8\linewidth]{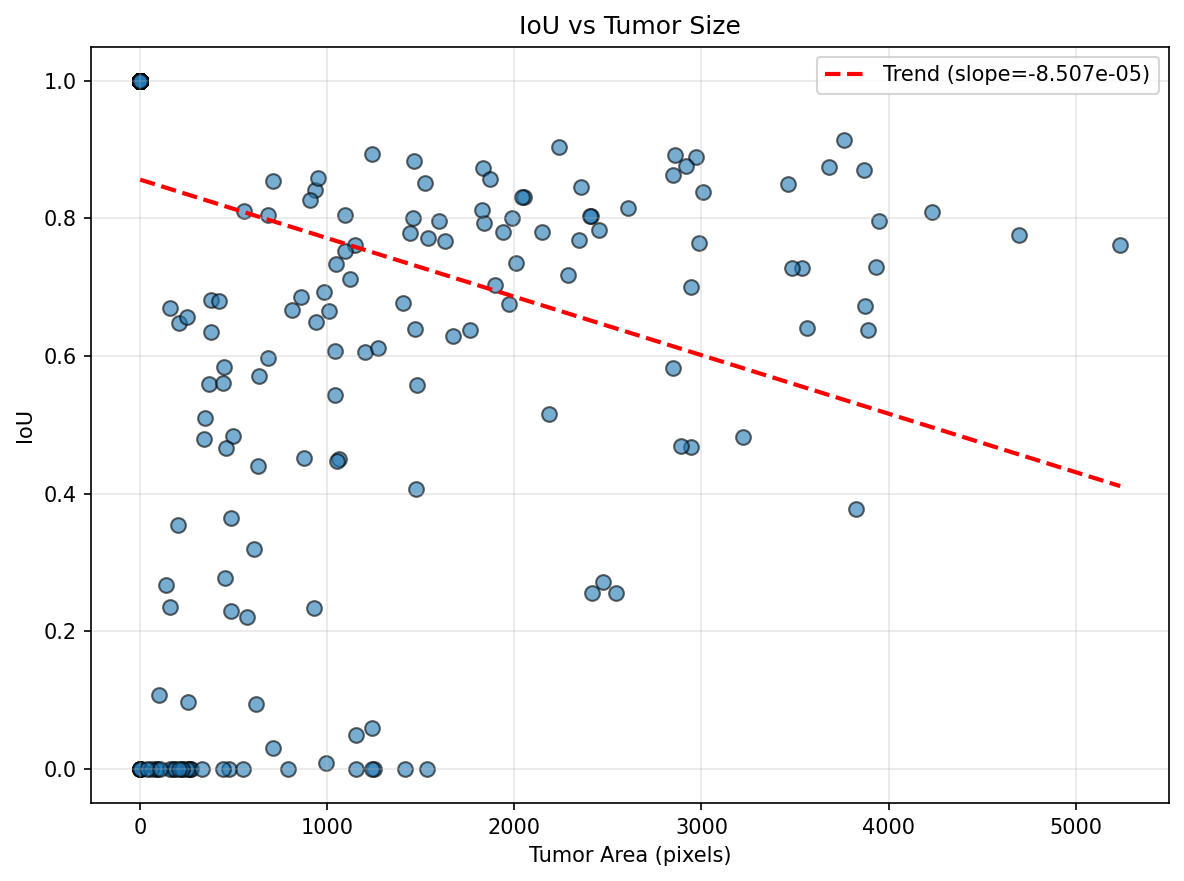}
\caption{IoU versus tumor area with a linear trend line.}
\label{fig:iou_size}
\end{figure}

Figure~\ref{fig:iou_size} shows the correlation between tumor size and IoU. This is seen as a moderate positive relationship ($r=0.37$): In general, segmentation accuracy increases with larger size of the tumours. However, the proposed model can detect many small tumors with satisfactory IoU, indicating the capacity to detect tumor with different size.
\subsection{Distribution Analysis}
\begin{figure}[htbp]
\centering
\includegraphics[width=0.9\linewidth]{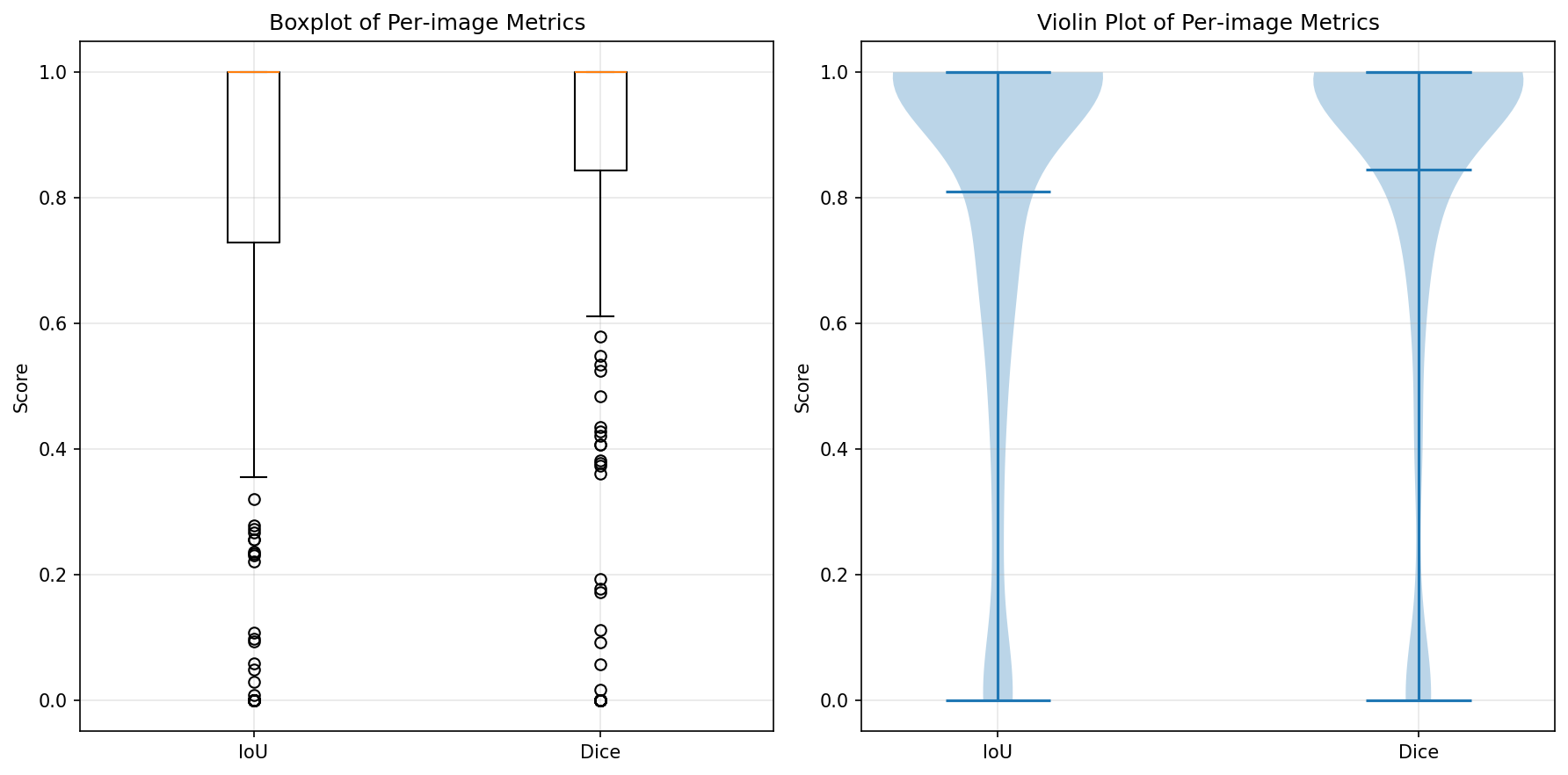}
\caption{Distribution of per-image IoU and Dice scores using box and violin plots.}
\label{fig:dist}
\end{figure}

Figure~\ref{fig:dist} Examples of the results are shown in per-image IoU and Dice scores. Results show consistently good segmentation performance, with median IoU (Dice) scores of 0.74 (0.99). Although the accuracy is high in most images, a few more difficult images show lower scores, which may be attributed to the differences in the size, shape and contrast of the images of tumors.
\subsection{Qualitative Results}
\begin{figure}[htbp]
\centering
\includegraphics[width=0.8\linewidth]{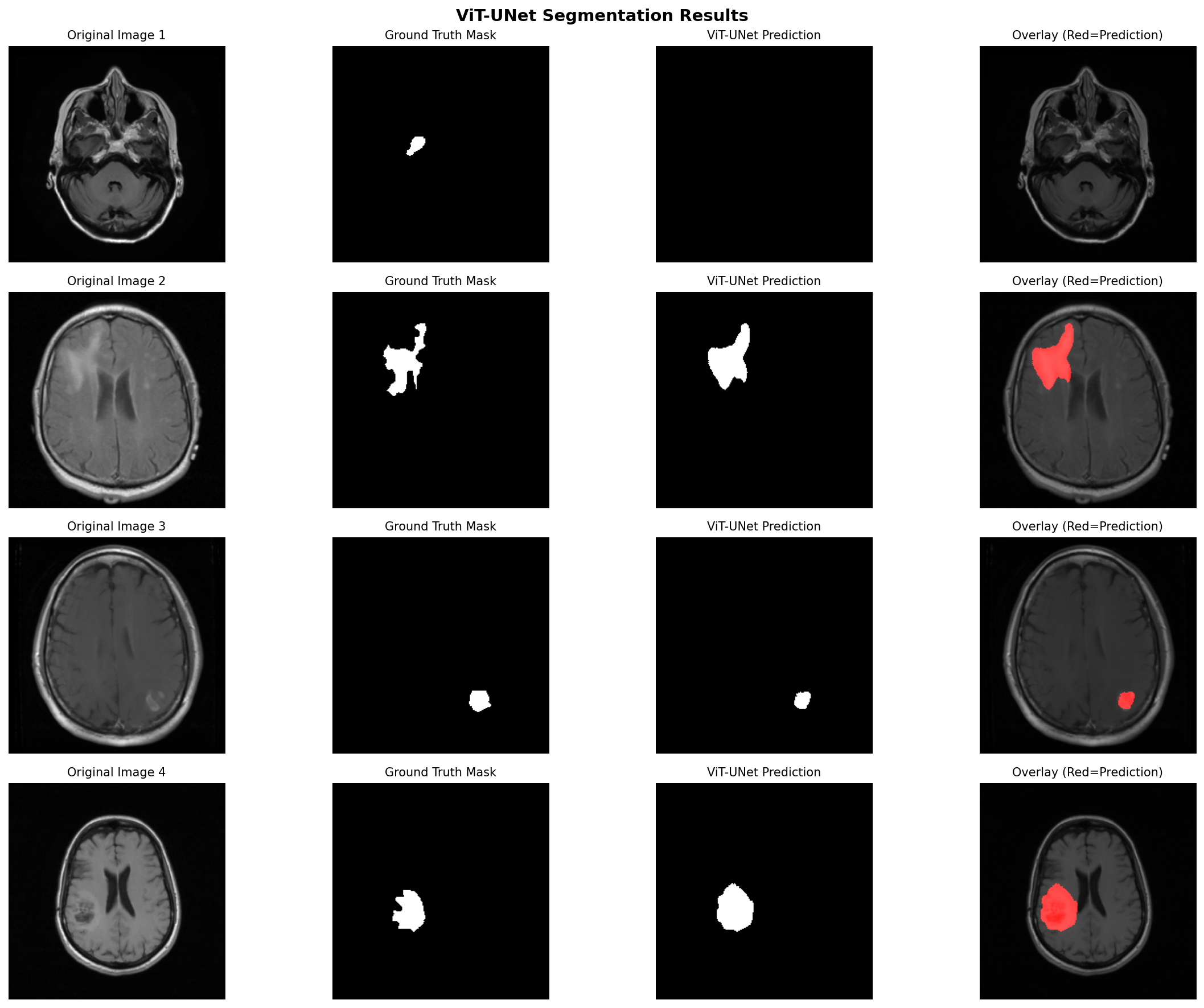}
\caption{Qualitative segmentation results: original image, ground truth, prediction, overlay.}
\label{fig:seg_results}
\end{figure}

\begin{figure}[htbp]
\centering
\includegraphics[width=0.7\linewidth]{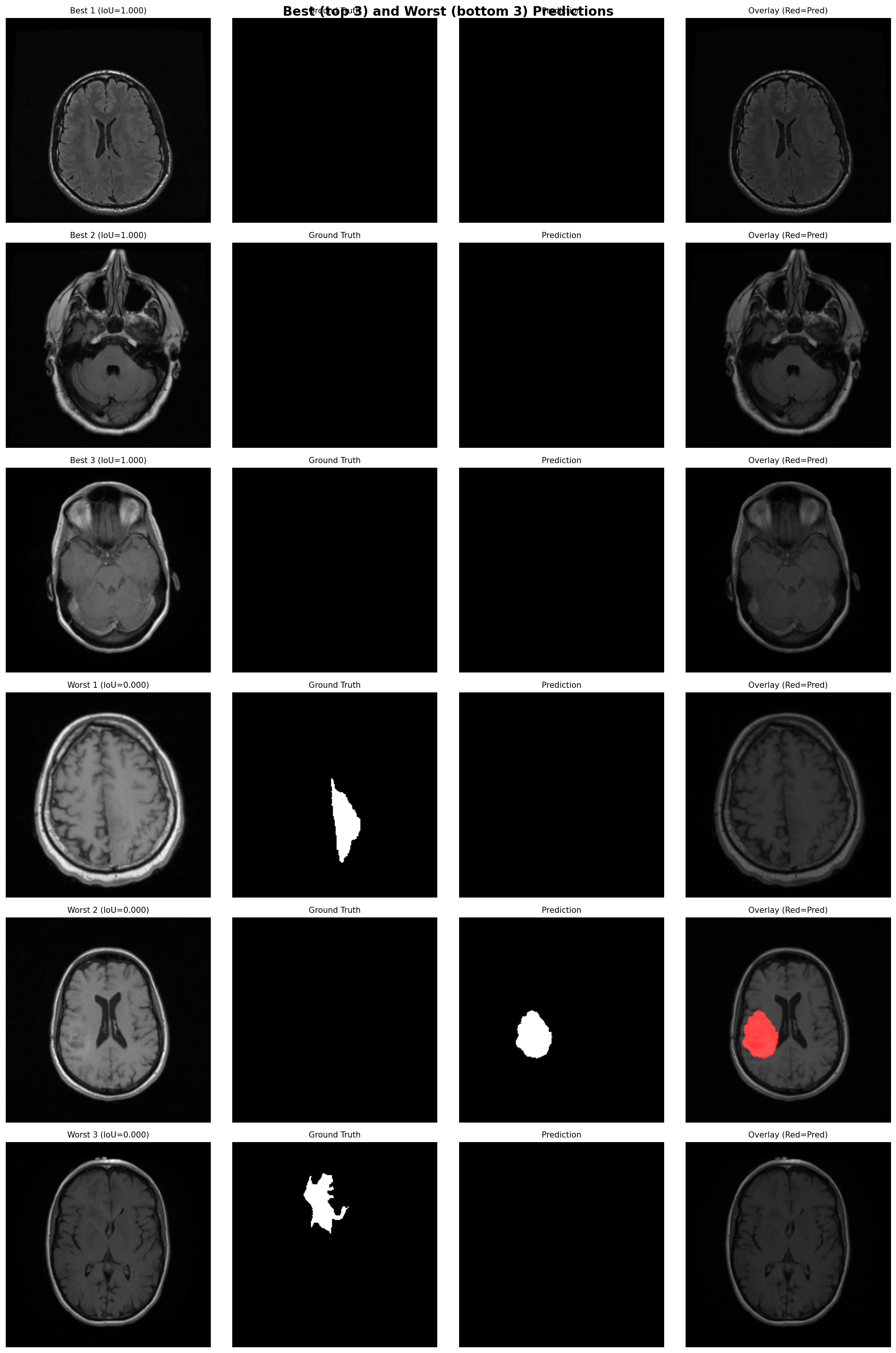}
\caption{Best and worst segmentation predictions based on IoU score, illustrating the range of performance.}
\label{fig:best_worst}
\end{figure}

Figure~\ref{fig:seg_results} Shows representative segmentation results. The proposed model is able to capture boundaries of tumors with high accuracy compared with the ground truth with irregular boundaries, with only minor discrepancies in the boundaries in challenging cases.
Additionally, we showcase the best and worst predicted cases in Figure~\ref{fig:best_worst} to highlight the model's strengths and failure modes.
\subsection{ABLATION STUDY}
To assess the impact of each of the parts of the proposed ViT-UNet architecture, an ablation study is performed on the LGG dataset, and the results are shown in Table IV. We successively examine the Vision Transformer (ViT) bottleneck and the hybrid loss function.
\begin{table}[t]
\caption{Ablation study of the proposed ViT-UNet architecture on the LGG dataset.}
\label{tab:ablation}
\centering
\begin{tabular}{lcccc}
\toprule
\textbf{Model Variant} & \textbf{ViT} & \textbf{Hybrid Loss} & \textbf{IoU} & \textbf{Dice} \\
\midrule
U-Net Baseline               & $\times$ & $\times$ & 0.7807 & 0.8188 \\
U-Net + Hybrid Loss          & $\times$ & $\checkmark$ & 0.7924 & 0.8296 \\
U-Net + ViT Bottleneck       & $\checkmark$ & $\times$ & 0.8031 & 0.8372 \\
\textbf{Proposed ViT-UNet}   & $\checkmark$ & $\checkmark$ & \textbf{0.8100} & \textbf{0.8446} \\
\bottomrule
\end{tabular}
\end{table}
\subsection{Training Dynamics}
\begin{figure}[htbp]
\centering
\includegraphics[width=0.9\linewidth]{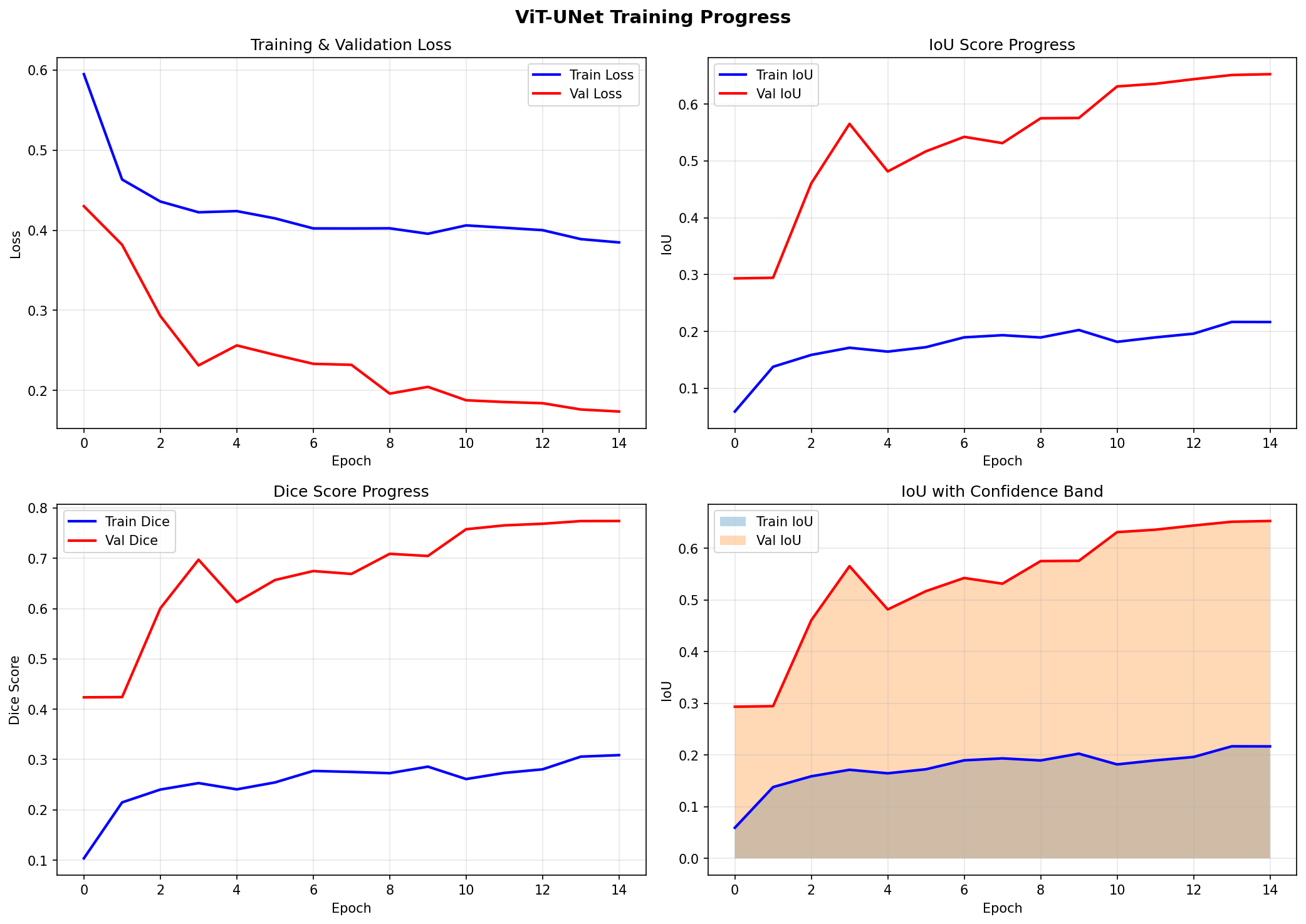}
\caption{Training and validation loss, IoU, and Dice over the training epochs.}
\label{fig:training}
\end{figure}

Figure~\ref{fig:training} shows the training process of the proposed model. The training and validation losses show a clear steady decline whereas IoU and Dice metrics show gradual improvement which illustrates the smooth convergence without overtraining. The cosine annealing learning rate schedule is also used which further helps to optimize the model during the training process.
\section{Discussion}
Evaluation of the proposed architecture, interpretation of the experiment results and suggestions for future research and limitations of this study are addressed.
\subsection{Performance Interpretation}
The proposed ViT-UNet achieved a mean IoU of 0.8100 and a Dice score of 0.8446, demonstrating promising brain tumor segmentation performance. The model achieved high specificity (99.73\%) and a sensitivity of 82.6\%, indicating effective tumor detection while maintaining a low false-positive rate. Further investigation will focus on multimodal MRI data and improved learning strategies to enhance segmentation performance and generalizability.
\subsection{Limitations and Future Work}
Despite the promising performance of the proposed model, this study has several limitations. The model was evaluated on a relatively small LGG dataset using single-modality 2D MRI images, which may limit its generalizability across different datasets, MRI modalities, and acquisition protocols. Future work will focus on evaluation using larger and more diverse datasets, multimodal and 3D segmentation, explainable AI, and real-time optimization to further improve the model's robustness and practical applicability.
\section{Conclusion}
This paper presented a lightweight brain tumor segmentation model based on a Vision Transformer (ViT) bottleneck incorporated into a U-Net architecture. The proposed model combines CNN-based local feature extraction with Transformer-based global contextual modeling while maintaining a compact size of approximately 2.6 million parameters. Experiments on the TCGA-LGG MRI segmentation dataset achieved an IoU of 0.8100 and a Dice score of 0.8446, outperforming the baseline U-Net. These results demonstrate the potential of the proposed architecture for accurate and computationally efficient brain tumor segmentation. Future work will investigate multimodal and 3D MRI segmentation, model explainability, and validation on larger and more diverse datasets.

\end{document}